\documentclass{article}

\usepackage{microtype}
\usepackage{graphicx}
\usepackage{booktabs} 
\usepackage{xcolor}
\usepackage{mathtools}
\usepackage{bbm}
\usepackage{subcaption}

\usepackage[normalem]{ulem}

\usepackage{pgfplots}
\usepgfplotslibrary{colorbrewer}

\usepackage{listings}
\usepackage{xcolor}
 
\definecolor{codegreen}{rgb}{0,0.6,0}
\definecolor{codegray}{rgb}{0.5,0.5,0.5}
\definecolor{codepurple}{rgb}{0.58,0,0.82}
\definecolor{backcolour}{rgb}{0.95,0.95,0.92}

\lstdefinestyle{mystyle}{
    backgroundcolor=\color{backcolour},   
    commentstyle=\color{codegreen},
    keywordstyle=\color{magenta},
    numberstyle=\tiny\color{codegray},
    stringstyle=\color{codepurple},
    basicstyle=\ttfamily\footnotesize,
    breakatwhitespace=false,         
    breaklines=true,                 
    captionpos=b,                    
    keepspaces=true,                 
    numbers=left,                    
    numbersep=5pt,                  
    showspaces=false,                
    showstringspaces=false,
    showtabs=false,                  
    tabsize=2,
    mathescape=true
}

\usepackage{xspace}

\pgfplotscreateplotcyclelist{mycolorlist}{%
blue!90!black,smooth,very thick\\%
red!80!black,smooth,very thick\\%
orange!95!black,smooth,very thick\\%
lime!90!black,smooth,very thick\\%
}
\pgfplotsset{
compat=1.12
}

\newcommand{\graph}[2][12]{
\begin{tikzpicture}
\begin{axis}[ymax=#1,ymin=7,xmax=2.5,cycle multi list=mycolorlist,
    xlabel={Training Time (days)},
    ylabel={Validation Perplexity},
    ymajorgrids=true,
    grid style=dashed,
    legend pos=north east,
    legend cell align={left},
    legend style={cells={align=left}},
    every axis plot/.append style={very thick}]
#2
\end{axis}
\end{tikzpicture}
}

\usepackage{hyperref}

\usepackage[accepted]{icml2021}

\begin{document}

\twocolumn[
\icmltitle{BASE Layers: Simplifying Training of Large, Sparse Models}



\icmlsetsymbol{equal}{*}

\begin{icmlauthorlist}
\icmlauthor{Mike Lewis}{fair}
\icmlauthor{Shruti Bhosale}{fair}
\icmlauthor{Tim Dettmers}{fair,uw}
\icmlauthor{Naman Goyal}{fair}
\icmlauthor{Luke Zettlemoyer}{fair,uw}
\end{icmlauthorlist}
\icmlaffiliation{fair}{Facebook AI Research}
\icmlaffiliation{uw}{University of Washington}

\icmlcorrespondingauthor{Mike Lewis}{mikelewis@fb.com}

\icmlkeywords{Machine Learning, ICML}

\vskip 0.3in
]



\printAffiliationsAndNotice{}  

\begin{abstract}

We introduce a new balanced assignment of experts (BASE) layer for large language models that greatly simplifies existing high capacity sparse layers. 
Sparse layers can dramatically improve the efficiency of training and inference by routing each token to specialized expert modules that contain only a small fraction of the model parameters. 
However, it can be difficult to learn balanced routing functions that make full use of the available experts; existing approaches typically use routing heuristics or auxiliary expert-balancing loss functions.
In contrast, we formulate token-to-expert allocation as a linear assignment problem, allowing an optimal assignment in which each expert receives an equal number of tokens.  
This optimal assignment scheme improves efficiency by guaranteeing balanced compute loads, and also simplifies training by not requiring any new hyperparameters or auxiliary losses. Code is publicly released.\footnote{\url{https://github.com/pytorch/fairseq/}}

\end{abstract}

\section{Introduction}
\label{submission}
Sparse expert models enable sparse computation by spreading model capacity across a set of experts, while ensuring that only a small subset of the experts are used for each input~\cite{moe,gshard,switch}. Sparse models can often realize the strong performance gains that come with training very large models, while also alleviating much of the associated computational, financial and environmental costs~\cite{strubell2019energy}. However, such models are notoriously difficult to train; the experts must be carefully balanced so that they can specialize to different parts of the input space. In this paper, we present a simple, efficient, and performant method for expert-based sparsity in language models, built around the use of a linear assignment algorithm to explicitly balance the assignment of tokens to experts during training. 

The mostly widely used Sparse Expert models are mixtures of experts (MoE) models~\cite{moe,gshard} that learn a gating function to route each token to a few experts, which creates a challenging, discrete latent variable learning problem. In practice, carefully tuning and the introduction of extra loss functions with new hyperparameters is required to avoid imbalanced or degenerate experts. Recently, the Switch transformer~\cite{switch} simplified the framework by routing tokens to only a single expert, improving stability and efficiency overall but again using custom auxiliary losses that require tuning, and requiring capacity factors to prevent too many tokens being assigned to a single expert. We show that it is possible to go even further. We also assign a single expert per token but are the first to algorithmically balance the assignment with no extra model modifications, providing more formal guarantees of balanced compute while simplifying both the implementation and optimization.

\begin{figure}
    \centering
    \includegraphics[width=\columnwidth]{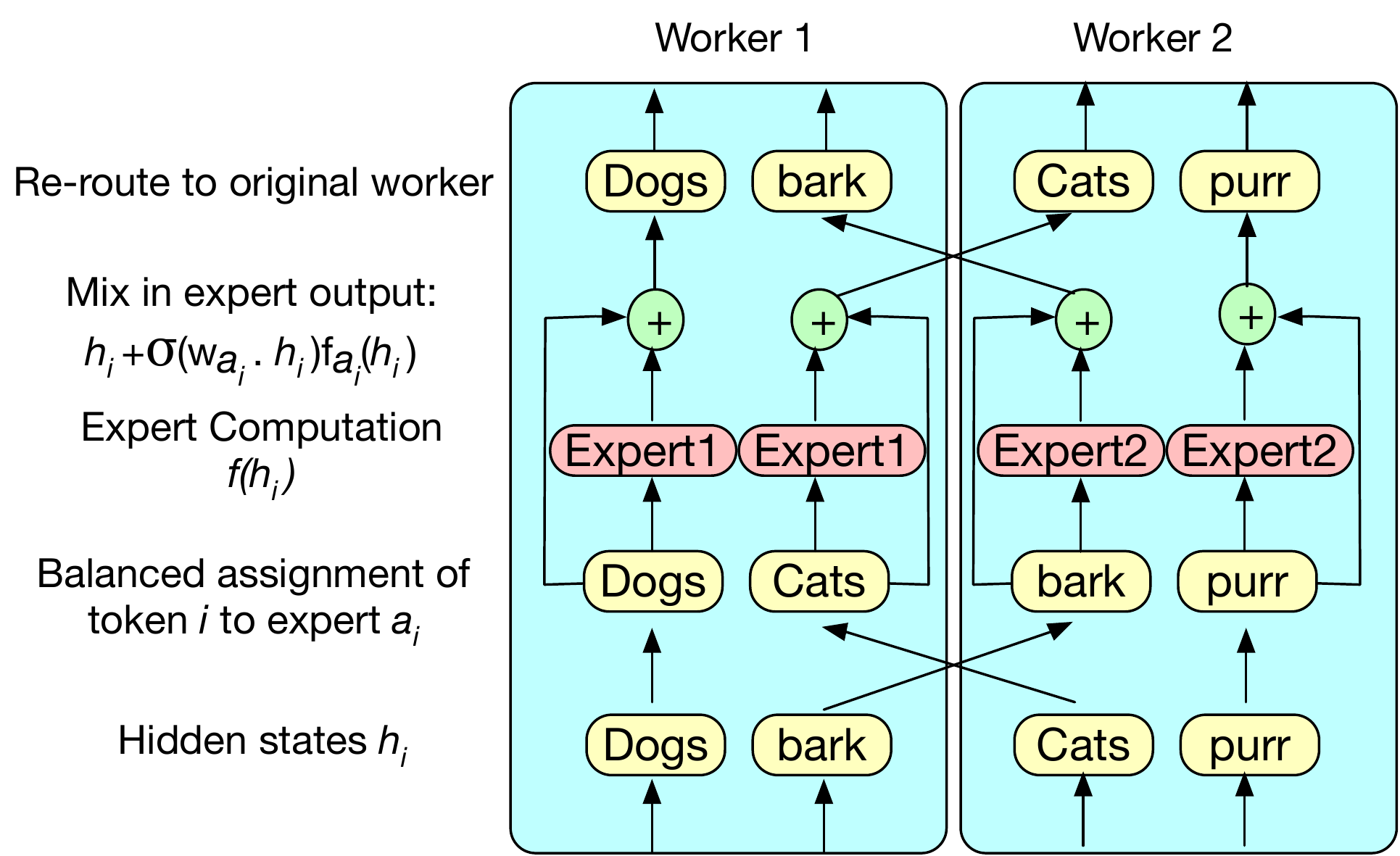}
    \caption{Overview of a BASE layer. Each worker contains a separate \emph{expert} module. During training, we compute a balanced assignment of tokens such that each worker sends an equal number of tokens to each expert. By softly mixing in the expert module, experts can learn to specialize for particular types of tokens. }
    \label{fig:overview}
\end{figure}

We introduce a simple and effective solution for routing tokens to experts during training, which we use to estimate a new Balanced Assignment of Sparse Experts (BASE) layer. 
To ensure balanced routing in the BASE layer, we formulate a linear assignment problem that maximizes token-expert affinities while ensuring that each expert receives an equal number of tokens.
This approach ensures that the assignment will be balanced, and therefore each expert will operate at maximum capacity, while also eliminating load-balancing loss functions and capacity factors from previous work. 
We also show how to learn expert specialization by using a modified residual connection that softly mixes in each expert contribution---again without requiring an additional loss term or routing tokens to multiple experts.
While computing balanced assignments incurs non-trivial overhead, we find that using even a single large BASE layer is remarkably effective---reduced expert communication produces faster gradient computations---and that performance increases as more BASE layers are added, providing an overall favorable cost-accuracy tradeoff.

Extensive experiments with models of up to 110B parameters demonstrate large performance gains over standard data and model parallel training strategies. Our approach also matches or exceeds the efficiency and performance of previous sparse expert approaches~\citep{gshard,switch}, when controlling for computation budget, despite its relative simplicity. Taken together, these results  demonstrate the first drop-in conditional compute layer that can be easily added to any model with no new hyperparameters or training loss modifications.

\section{Background: Training with Multiple Workers}

NLP has recently become dominated by ever larger language models \cite{bert,bart,roberta,gpt2,t5}.
Training large language models would take infeasibly long on any existing single device, with many models trained for thousands of GPU-days \cite{gpt3}. Instead, it is standard to distribute computation over multiple workers. We briefly review the main existing strategies.

\subsection{Dense Models}
In dense models, every parameter is used in processing every input. Training is distributed over multiple workers using data parallism or model parallelism.

\paragraph{Data Parallel Training} In data parallel training, multiple workers maintain a copy of the same model. Each worker runs the model on a different subset of the training batch, then gradients are communicated and all workers perform the same update. This approach increases the number of examples processed per second, and only requires a single communication step between workers per update. However, the maximum model size that can be trained is bounded by the memory of a single worker device---limiting models to roughly 1.5B parameters in our setup.

\paragraph{Model Parallel Training} Model parallel training allows models to be larger than can be run on a single worker \cite{megatron}, by distributing the compute for each input over multiple workers. Model parameters are also distributed over workers, which then communicate with each other while processing each input. Given a fixed number of workers, using model parallel training will reduce the amount of compute available for data parallelism, and correspondingly also the number of examples processed per second.

\subsection{Sparse Expert Layers}
Sparse models differ from dense models in only using a small subset of their parameters on any given input.
Recent work has explored adding capacity to language models by adding sparse expert layers \cite{moe,gshard,switch}. During inference, before an expert layer, each token is assigned and routed to a small subset of the workers. The workers then applies a token-wise operation, using parameters that are not shared across other workers. The resulting representation is then returned to the original worker, to continue the forward pass.

During training, this results in four routing steps per expert layer---before and after each expert layer, in both the forward and backward pass. These communication steps can add significantly to the training cost, as workers can idle while waiting for communication to complete.

Balancing of experts, so that each  processes a roughly equal proportion of tokens, is crucial for several reasons. If one expert is assigned too many tokens, the worker could run out of memory. Additionally, the expert layer processing speed is limited by the slowest worker; imbalanced assignment slows down training. Furthermore, the parameters of rarely used experts are likely to be less well trained, which may reduce performance. 

Previous work has achieved balancing by adding a new term in the loss function that explicitly encourages balancing---this loss term must be carefully weighted so that it does not overwhelm primary loss~\cite{gshard,switch}. However, such a loss does not guarantee balancing. Stable training also requires additional measures such as enforcing hard upper limits on the number of tokens processed by each expert after which the rest are simply ignored \cite{moe}. 
This approach can be inefficient, as some workers are underutilized, and many tokens are unprocessed by the layer.

\begin{figure*}
    \centering
    \input{pseudocode}
    \caption{ Implementation of a BASE layer, with $E$ experts and an input sequence of $T$ features. Here, \emph{all\_to\_all} routes the $t$th row of its input to the $\lfloor \frac{tE}{T}\rfloor$th worker. \emph{balanced\_assignment} takes a matrix of size $T\times E$ and returns an $T$-dimensional vector that can be used to sort tokens by their assigned expert index. 
    }
    \label{fig:pseudocode}
\end{figure*}

\section{BASE Layers}
\label{section:base}

BASE layers achieve balanced assignment of tokens to experts through a three stage process. Firstly, we compute the score for assigning each token representation to each expert, compute a balanced assignment maximizing these scores, then route the token features to an expert. Secondly, we compute a position-wise expert function, and compute a weighted sum of the layers input and output. Finally, we return the output to the original worker. Figure~\ref{fig:pseudocode} shows overall pseudo code for the approach.


\subsection{Parameterization}
BASE layers contain $E$ experts, each defined by a position-wise function $f_e(\cdot)$ and an expert embedding $w_e \in R^D$, where $D$ is the model dimension. In practice, we parameterize $f_e(\cdot)$ using a stack of residual feedforward layers. Given a token $h_t$ at timestep $t$ in a sequence of tokens $0..T$, and token-to-expert assignment index $a_t\in 0..E$, the network returns the following value: 
\begin{equation}
    \sigma(h_t\cdot w_{a_t})f_{a_t}(h_t) + h_t, 
\end{equation}
If the network $f_{a_t}$ is able to improve the representation of $h_t$, by lowering the loss of the final prediction for that token, then gradient descent will increase the value of $h_t\cdot w_{a_t}$. Conversely, if the expert network is unhelpful, then the $h_t\cdot w_{a_t}$ will receive a negative gradient. Consequently, an expert $e$ can learn to specialize for particular types of tokens by adjusting $w_e$ to be close to similar token representations where $f_e(\cdot)$ is most beneficial.

\subsection{Token to Expert Assignment}
We assign tokens to experts using different methods during training and testing. During training, we maximize model throughput by assigning an equal number of tokens to each expert. At test time, we simply assign each token to its highest scoring expert.

\subsubsection{Assignment During Training}
\label{section:routing_training}
During training, we assign an equal number of tokens to each expert, so that each worker is fully utilized and each worker takes about the same time to finish its assigned load.


Each token $t$ is assigned to an expert $a_t$, aiming to maximize the token-expert affinities under the constraints that each expert is assigned the same number of tokens.

\paragraph{Linear Assignment Problem}
Formally, we solve the following linear assignment problem. Given $T$ tokens with representations $h_t$ and $E$ experts with embeddings $w_e$, we assign each token to an expert via the assignment index $a_t\in 0..E$:
\begin{align}
\begin{split}
    \text{maximize} \sum_t h_{t}\cdot w_{a_t} \\
    \text{subject to } \forall e \sum_{t=0}^{T} \mathbbm{1}_{a_t=e}=\frac{T}{E} 
\end{split}
\end{align}

Numerous algorithms exist for this problem. We use the auction algorithm described in \citet{auction}, which is more easily parallelizable on GPUs than the Hungarian Algorithm \cite{hungarian}. Pseudo-code is given in the Appendix.

\paragraph{Sharding} Computing the optimal assignment for all tokens across all workers is expensive, so we distribute the computation across multiple workers. We decompose the assignment problem of all $ET$ tokens across all workers into $E$ smaller problems using $T$ tokens. This decomposition can be implemented by each worker solving an assignment problem over its own input batch. Each worker then sends $T/E$ tokens to each other worker, with an \emph{all2all} operation.

\paragraph{Shuffling}
Tokens within each worker's training sequence are highly correlated with each other; for example they will normally be part of the same domain. These correlations may make it difficult for experts to specialize for particular domains. We therefore add an additional \emph{random} routing step, where each worker first sends an equal number of each tokens to each other worker \emph{randomly}. Then, each worker solves a linear assignment problem as before with its sample of tokens, and routes these to the correct experts. 




\subsubsection{Assignment During Testing}
At test time, it is not possible to use the assignment strategy described in \S\ref{section:routing_training}, as balancing the assignment leaks information about tokens in the future context. Instead, we simply greedily assign the one best expert. While unbalanced assignments are less efficient, during inference memory costs are greatly reduced due to not needing to store gradients, activations and optimizer states. In practice, we show that our approach naturally learns a reasonably balanced assignment during training (\S\ref{section:balance}).



\subsection{Gradient Clipping}
A common practice in training deep language models is to scale gradients if their $l_2$ norm is greater than a threshold. 
All workers must compute the same norm, or else scaled gradients for shared parameters will be inconsistent across workers.
To avoid additional communication steps to compute norms globally across all expert parameters, we simply compute the gradient norms locally based only on the shared parameters, but rescale all gradients.

\section{Experiments}
\label{section:results}
\subsection{Experimental Setup}
\label{section:exp_setup}
\paragraph{Task} We focus our experiments on language modelling, as recent work such as GPT3 \cite{gpt3} offers perhaps the clearest demonstration in machine learning of the power of large scale models. 

\paragraph{Metrics}
We focus exclusively on comparing \emph{compute efficiency}, which we define as the best model performance (here, perplexity) that can be achieved by training with a given number of GPUs and wall-clock time. This metric is different from other commonly used metrics, such as \emph{sample efficiency} (which measures the number of tokens the model trains on, but not the cost of processing samples) or \emph{FLOP-efficiency} (which measures the number of floating-point operations performed during training, but does not account for communication costs). As plentiful data is available for training language models, but computation is expensive, we believe that compute efficiency best captures the constraints of real world training. Therefore, we compare models using a fixed number of GPUs for the same runtime.

\paragraph{Training Hyperparameters}
We train all models for approximately 2.5 days. All models use similar hyperparameters of 2000 warm-up steps, and the Adam optimizer \cite{adam}. We tune learning rates for each model separately, and linearly decay the learning rate during training.  Each worker processes two sequences of length 1024, and gradients are accumulated over 8 updates. We clip gradients if their $l_2$ norm exceeds 0.1 (\S\ref{section:base}). Learning rates are tuned in the range $\{0.5, 0.75, 1.0\}\times 10^{-4}$, taking the highest value that avoids divergence.

\paragraph{Hardware}
Unless otherwise stated, models are trained on 128 32GB V100 GPUs connected with Infiniband.\footnote{As communication between workers is a significant overhead for model parallel and sparse expert approaches, it is possible that different results would be achieved on other networking hardware.} 
\begin{figure*}
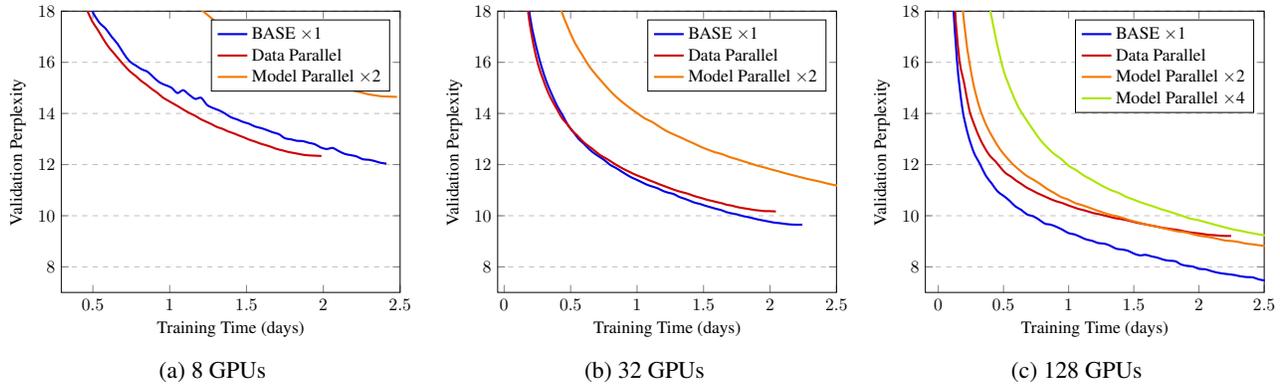

\begin{subfigure}{.33\textwidth}
\resizebox{\textwidth}{!}{%
    \graph[18]{\input{plots/8gpu/base}\input{plots/8gpu/model_parallel1}\input{plots/8gpu/model_parallel2}\legend{BASE $\times$1,Data Parallel, Model Parallel $\times$2}}
}
\caption{8 GPUs}
\end{subfigure}
\begin{subfigure}{.33\textwidth}
\resizebox{\textwidth}{!}{    \graph[18]{\input{plots/32gpu/base}\input{plots/32gpu/model_parallel1}\input{plots/32gpu/model_parallel2}\legend{BASE $\times$1,Data Parallel, Model Parallel $\times$2}}
}
\caption{32 GPUs}
\end{subfigure}
\begin{subfigure}{.33\textwidth}
\resizebox{\textwidth}{!}{%
    \graph[18]{\input{plots/moe_central}\input{plots/model_parallel1}\input{plots/model_parallel2}\input{plots/model_parallel4}\legend{BASE $\times$1,Data Parallel, Model Parallel $\times$2, Model Parallel $\times$4}}
}    
\caption{128 GPUs}
\end{subfigure}

    \caption{Comparing BASE layers with dense model training, using different numbers of GPUs.
    There is a clear trend of increased model sizes being more effective with larger compute budgets. BASE layers show strong performance at all the compute budgets we consider. }
    \label{fig:vs_baseline}
\end{figure*}
\paragraph{Data}
We train on a corpus of approximately 100B tokens, comprising the training corpus of RoBERTa \cite{roberta}, combined with the English portion of the CC100 corpus \cite{xlmr}.
We use the byte-pair encoding \cite{bpe} from GPT2 \cite{gpt2}, which has a vocabulary of 51200.

\paragraph{Model Architectures}
We size all models to the maximum size that can process the sequences within GPU memory constraints. All models follow a standard transformer architecture \cite{transformer}, with a model dimension of 2048, feed-forward hidden states of size 8096 and 24 Transformer Decoder blocks. We use 16 attention heads, ReLU activation functions and no dropout. LayerNorm \cite{layernorm} is applied to the inputs of each residual block \cite{prenorm} and to the outputs of the transformer. 

\paragraph{BASE layer architecture} We implement the BASE layer as a stack of feedforward blocks. Each block follows the standard transformer structure: layer normalization, a projection to 4 times the input dimension, a ReLU nonlinearity, a projection to the input dimension, and a residual connection to the block input.
We vary the number of BASE layers; BASE$\times N$ uses a BASE layer after each of the $\lfloor\frac{L}{N+1}\rfloor\dots\lfloor\frac{NL}{N+1}\rfloor$th transformer layers.
When using multiple BASE layers, we reduce their size to keep the total number of parameters roughly constant; BASE$\times N$ use $\lfloor \frac{10}{N}\rfloor$ sublayers, for a total of roughly 44B parameters.
We use one expert per GPU per BASE layer.

\subsection{Comparison with Dense Models}
We first compare with dense models, in which all parameters are shared across all workers. We compare with data parallel and model parallel training, using the intra-layer model parallelism approach introduced in \citet{megatron}. Our data parallel baseline contains 1.5B parameters, and the 2-way and 4-way model parallel baselines contain roughly 3B and 6B parameters respectively.
We use three different compute budgets: 8, 32 and 128 GPUs for approximately 2.5 days. 

Results are shown in Figure \ref{fig:vs_baseline}. We generally find that larger models perform better with higher compute budgets, and that simple data parallel training performs best at the smallest compute budget. With larger compute budgets, BASE layers outperform both data parallel and model parallel training by a wide margin.

Relatively high compute budgets are required before model parallelism outperforms data parallel training, with the first gains appearing after training on 128 GPUs for 2 days. This is partly due to model parallel training requiring a reduced batch size given the same computational resources.

In contrast, BASE layers match the performance of data parallel training on our 8 GPU experiments, and achieve increasingly large gains in higher compute regimes.

\subsection{Comparison with Sparse Experts Models}

We also compare performance with our re-implementations of two recent sparse layer methods: Sparsely Gated Mixtures of Experts \cite{moe,gshard} and Switch \cite{switch}. The primary difference between these approaches is that a Sparsely Gated MoE layer routes tokens to multiple experts (top-2 experts in our experiments), whereas a Switch layer routes tokens to a single expert. We set the weight associated with the load balancing loss to 0.01 in our experiments, and set the capacity factor for Sparsely Gated MoE and Switch layers to 2.0 and 1.0 respectively. Following previous work, we replace every other shared feed-forward layer in the Transformer architecture with a Sparsely Gated MoE or Switch layer, unless otherwise specified. With 128 experts in each expert layer, our Sparsely Gated MoE and Switch models have 52.5B parameters (1B shared parameters) each, while our BASE model has 44.4B parameters (1.3B shared parameters).

\begin{figure}[t]
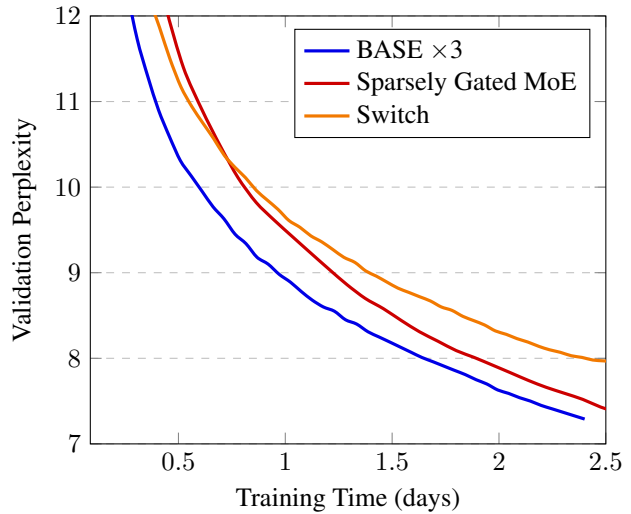

    \centering
    \graph{\input{plots/moe_3layer}\input{plots/gshard}\input{plots/switch}\legend{BASE $\times$3,Sparsely Gated MoE, Switch}}
    \caption{Comparison with other Sparse Experts approaches. Despite its simplicity, BASE achieves strong performance relative to Sparsely Gated MoE models and Switch transformers.}
    \label{fig:vs_google} 
\end{figure}

As in \citet{switch}, we find that Switch computes more updates per second than Sparsely Gated MoE (see \autoref{tab:efficiency}). However, we find that Sparsely Gated MoE is more compute efficient in our experiments as shown in \autoref{fig:vs_google}.

A comparison with BASE is also shown in \autoref{fig:vs_google}. 
Despite its simplicity, BASE achieves similar performance to the Sparsely Gated MoE model and converges to a better validation perplexity than Switch. This result suggests that algorithmic load balancing is a competitive alternative to load balancing loss functions, and that even a single expert layer can be highly effective.

\subsection{Ablations}
\label{section:ablations}
Results in Section \ref{section:results} show that BASE layers match or exceed the compute-efficiency of previous dense and sparse approaches. To better understand these results, we analyze key design decisions in our model in more detail.

\paragraph{BASE Layer Size}
A key choice in any sparse experts model is the allocation of capacity to shared components versus experts. We experiment with adjusting the number of sublayers in each expert, and scale the number of shared layers accordingly to maximize GPU usage.

We test three versions:
\begin{itemize}
    \item \textbf{Small Expert}: 1.5B shared parameters, 135M parameters per expert, 18.8B total parameters
    \item \textbf{Standard Expert}: 1.3B shared parameters, 335M parameters per expert, 44B total parameters
    \item \textbf{Large Expert}: 840M shared parameters, 911M parameters per expert, 117B total parameters
\end{itemize}

\begin{figure}[t]
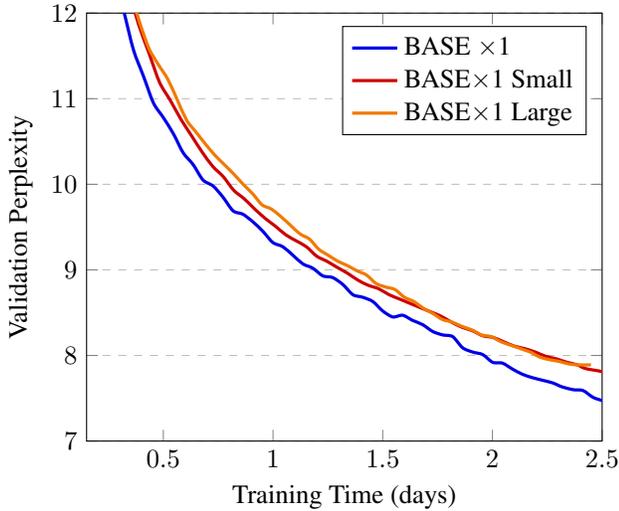

    \centering
\graph{\input{plots/moe_central}\input{plots/moe_small}\input{plots/moe_big}\legend{BASE $\times$1,BASE$\times$1 Small,BASE$\times$1 Large}}
    \caption{Comparison of different sizes of BASE layers, by changing the ratio of parameters allocated to shared vs. expert layers.}
    \label{fig:expert_size}
\end{figure}

Figure \ref{fig:expert_size} shows that good performance can be achieved with all sizes, indicating that this choice needs little tuning.

\paragraph{BASE Layer Position}
We also consider the most effective place in a model to insert BASE layers into a transformer with $L$ layers. We test three configurations:
\begin{itemize}
\item \textbf{BASE}: After the $\frac{L}{2}$th layer, as in our other experiments.
\item \textbf{BASE Top}: After the $L$th layer, acting as a classifier.
\item \textbf{BASE $\times N$}: Using $N$ BASE layers of $\frac{1}{N}$ the size, after layers $\frac{L}{N+1}\dots\frac{NL}{N+1}$th layers of the transformer.
\end{itemize}

\begin{figure}
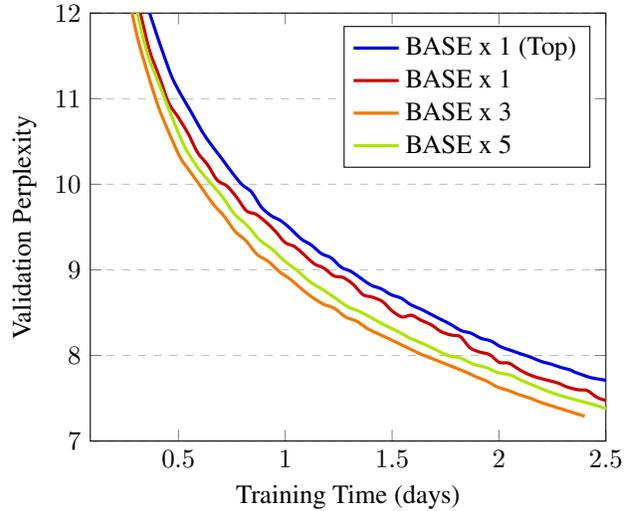

    \centering
    \graph{\input{plots/moe_top}\input{plots/moe_central}
    \input{plots/moe_3layer}\input{plots/moe_5layer}
    \legend{BASE x 1 (Top),BASE x 1,
    BASE x 3, BASE x 5
    }}
    \caption{Comparison of different numbers and positions of BASE layers. The best performance is achieved by interleaving 3 BASE layers throughout the transformer stack.  
    }
    \label{fig:expert_position}
\end{figure}

Figure \ref{fig:expert_position} compares results for different configurations. We find similar performance from three different placements of BASE, suggesting a reasonable level of robustness. In particular, the strong performance of BASE Top may enable it to be used on top of pre-trained language models to further increase their capacity.

\paragraph{Comparison of Routing Method with Sparsely Gated MoE }
Our approach differs from previous work on sparse experts in both the architecture and assignment method. To more carefully analyse the benefits of our routing method, we compare with an implementation of Sparsely Gated MoE  that uses a more similar architecture to ours: a single, large expert midway through the transformer stack. 

\begin{figure}
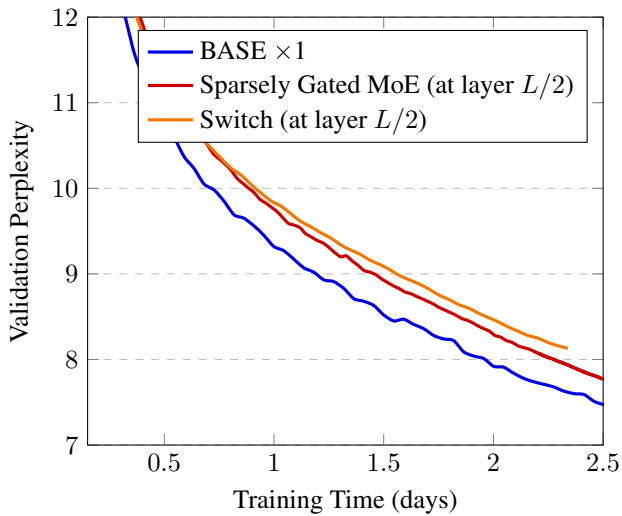

    \centering
    \graph{\input{plots/moe_central}\input{plots/gshard_central}\input{plots/switch_central}
    \legend{BASE $\times$1,Sparsely Gated MoE  (at layer $L/2$),Switch (at layer $L/2$)}}
    \caption{Comparing routing strategies using similar architectures. Here, all models use a single large expert at layer $L/2$. BASE maintains strong performance in this setting, which reduces the communication overhead between workers, and may be advantageous with less efficient networking. }
    \label{figure:gshard_mid}
\end{figure}

Results are shown in Figure \ref{figure:gshard_mid}. Sparsely Gated MoE  performs less well in this setting. Sparsely Gated MoE  benefits from interleaving expert layers with shared layers, and a single Sparsely Gated MoE layer with deep experts works less well than BASE. Future work should explore more efficient approximate routing schemes for BASE layers, to enable potential compute efficiency gains from interleaving expert and shared layers.

\section{Analysis}

We also report further experiments that provide more qualitative analyses of overall model behavior with BASE layers. 

\begin{figure}[t]
\begin{tikzpicture}
\definecolor{color1}{HTML}{440d54}
\definecolor{color2}{HTML}{277f8e}
\definecolor{color3}{HTML}{9fda3a}
\definecolor{color4}{HTML}{fde725}

\begin{axis}[
    xlabel={Experts (sorted by usage)},
    ylabel={Percentage of Tokens Routed to Expert},
    xmin=0, xmax=127,
    ymin=0, ymax=10,
    xtick={0,20,40,60,80,100,120},
    log ticks with fixed point,
    legend cell align={left},
    ymajorgrids=true,
    grid style=dashed,
    every axis plot/.append style={very thick}
]

\addplot[
    color=red!80!black,
    ]
    coordinates {
(0,1.76) (1,1.55) (2,1.32) (3,1.25) (4,1.19) (5,1.14) (6,1.10) (7,1.07) (8,1.05) (9,1.03) (10,1.02) (11,1.00) (12,0.99) (13,0.98) (14,0.97) (15,0.96) (16,0.95) (17,0.94) (18,0.94) (19,0.93) (20,0.92) (21,0.92) (22,0.91) (23,0.90) (24,0.90) (25,0.89) (26,0.89) (27,0.88) (28,0.88) (29,0.88) (30,0.87) (31,0.87) (32,0.86) (33,0.86) (34,0.86) (35,0.85) (36,0.85) (37,0.84) (38,0.84) (39,0.84) (40,0.83) (41,0.83) (42,0.83) (43,0.82) (44,0.82) (45,0.82) (46,0.81) (47,0.81) (48,0.81) (49,0.80) (50,0.80) (51,0.80) (52,0.80) (53,0.79) (54,0.79) (55,0.79) (56,0.78) (57,0.78) (58,0.78) (59,0.78) (60,0.77) (61,0.77) (62,0.77) (63,0.77) (64,0.76) (65,0.76) (66,0.76) (67,0.75) (68,0.75) (69,0.75) (70,0.75) (71,0.74) (72,0.74) (73,0.74) (74,0.74) (75,0.73) (76,0.73) (77,0.73) (78,0.73) (79,0.72) (80,0.72) (81,0.72) (82,0.71) (83,0.71) (84,0.71) (85,0.71) (86,0.70) (87,0.70) (88,0.70) (89,0.69) (90,0.69) (91,0.69) (92,0.69) (93,0.68) (94,0.68) (95,0.68) (96,0.67) (97,0.67) (98,0.67) (99,0.66) (100,0.66) (101,0.65) (102,0.65) (103,0.65) (104,0.64) (105,0.64) (106,0.63) (107,0.62) (108,0.62) (109,0.61) (110,0.60) (111,0.60) (112,0.59) (113,0.58) (114,0.57) (115,0.56) (116,0.55) (117,0.54) (118,0.52) (119,0.51) (120,0.50) (121,0.49) (122,0.48) (123,0.46) (124,0.45) (125,0.43) (126,0.41) (127,0.30)
    };
\addplot[
    color=orange!95!white,
    ]
    coordinates {
(0,9.17) (1,6.77) (2,4.85) (3,4.21) (4,3.74) (5,3.36) (6,3.03) (7,2.74) (8,2.55) (9,2.39) (10,2.26) (11,2.11) (12,1.98) (13,1.86) (14,1.75) (15,1.66) (16,1.57) (17,1.49) (18,1.42) (19,1.35) (20,1.28) (21,1.22) (22,1.16) (23,1.10) (24,1.05) (25,1.01) (26,0.97) (27,0.94) (28,0.90) (29,0.88) (30,0.85) (31,0.82) (32,0.79) (33,0.77) (34,0.74) (35,0.72) (36,0.70) (37,0.68) (38,0.65) (39,0.63) (40,0.61) (41,0.60) (42,0.58) (43,0.56) (44,0.55) (45,0.53) (46,0.52) (47,0.50) (48,0.49) (49,0.48) (50,0.46) (51,0.45) (52,0.44) (53,0.43) (54,0.42) (55,0.41) (56,0.40) (57,0.39) (58,0.38) (59,0.37) (60,0.36) (61,0.35) (62,0.34) (63,0.33) (64,0.33) (65,0.32) (66,0.31) (67,0.30) (68,0.30) (69,0.29) (70,0.28) (71,0.28) (72,0.27) (73,0.27) (74,0.26) (75,0.25) (76,0.25) (77,0.24) (78,0.24) (79,0.23) (80,0.23) (81,0.22) (82,0.22) (83,0.22) (84,0.21) (85,0.21) (86,0.20) (87,0.20) (88,0.20) (89,0.19) (90,0.19) (91,0.18) (92,0.18) (93,0.18) (94,0.17) (95,0.17) (96,0.16) (97,0.16) (98,0.16) (99,0.15) (100,0.15) (101,0.15) (102,0.14) (103,0.14) (104,0.14) (105,0.13) (106,0.13) (107,0.12) (108,0.12) (109,0.12) (110,0.11) (111,0.11) (112,0.11) (113,0.10) (114,0.10) (115,0.09) (116,0.09) (117,0.09) (118,0.08) (119,0.08) (120,0.07) (121,0.07) (122,0.06) (123,0.05) (124,0.05) (125,0.04) (126,0.03) (127,0.02)
    };
\addplot[
    color=lime!70!black,
    densely dashed,
    ultra thick,
]
    coordinates {
(0,1.74) (1,1.55) (2,1.39) (3,1.25) (4,1.21) (5,1.18) (6,1.15) (7,1.13) (8,1.11) (9,1.09) (10,1.08) (11,1.07) (12,1.05) (13,1.04) (14,1.03) (15,1.02) (16,1.01) (17,1.00) (18,0.99) (19,0.98) (20,0.97) (21,0.97) (22,0.96) (23,0.95) (24,0.94) (25,0.94) (26,0.93) (27,0.92) (28,0.92) (29,0.91) (30,0.90) (31,0.90) (32,0.89) (33,0.88) (34,0.88) (35,0.87) (36,0.87) (37,0.86) (38,0.86) (39,0.85) (40,0.85) (41,0.84) (42,0.84) (43,0.83) (44,0.83) (45,0.82) (46,0.82) (47,0.81) (48,0.81) (49,0.80) (50,0.80) (51,0.80) (52,0.79) (53,0.79) (54,0.78) (55,0.78) (56,0.77) (57,0.77) (58,0.77) (59,0.76) (60,0.76) (61,0.76) (62,0.75) (63,0.75) (64,0.75) (65,0.74) (66,0.74) (67,0.74) (68,0.73) (69,0.73) (70,0.73) (71,0.72) (72,0.72) (73,0.72) (74,0.71) (75,0.71) (76,0.71) (77,0.70) (78,0.70) (79,0.70) (80,0.69) (81,0.69) (82,0.69) (83,0.68) (84,0.68) (85,0.68) (86,0.68) (87,0.67) (88,0.67) (89,0.67) (90,0.66) (91,0.66) (92,0.66) (93,0.65) (94,0.65) (95,0.65) (96,0.64) (97,0.64) (98,0.63) (99,0.63) (100,0.63) (101,0.62) (102,0.62) (103,0.62) (104,0.61) (105,0.61) (106,0.60) (107,0.60) (108,0.59) (109,0.59) (110,0.58) (111,0.57) (112,0.56) (113,0.56) (114,0.55) (115,0.54) (116,0.53) (117,0.52) (118,0.51) (119,0.50) (120,0.49) (121,0.48) (122,0.46) (123,0.45) (124,0.42) (125,0.40) (126,0.37) (127,0.32)
    };
\addplot[
    color=cyan,
    ]
    coordinates {
    (0,0.78)(128,0.78)
    };
\addplot[
    color=blue!90!black,
]
    coordinates {
(0,4.05) (1,3.39) (2,2.74) (3,2.68) (4,2.42) (5,2.23) (6,2.21) (7,2.09) (8,2.07) (9,2.00) (10,1.95) (11,1.95) (12,1.81) (13,1.69) (14,1.64) (15,1.64) (16,1.61) (17,1.53) (18,1.52) (19,1.48) (20,1.45) (21,1.29) (22,1.24) (23,1.12) (24,1.08) (25,1.07) (26,1.04) (27,1.03) (28,0.99) (29,0.99) (30,0.97) (31,0.95) (32,0.94) (33,0.93) (34,0.93) (35,0.90) (36,0.90) (37,0.84) (38,0.82) (39,0.82) (40,0.81) (41,0.79) (42,0.79) (43,0.78) (44,0.77) (45,0.77) (46,0.74) (47,0.73) (48,0.71) (49,0.70) (50,0.67) (51,0.66) (52,0.66) (53,0.65) (54,0.61) (55,0.61) (56,0.60) (57,0.58) (58,0.57) (59,0.57) (60,0.57) (61,0.57) (62,0.57) (63,0.55) (64,0.55) (65,0.55) (66,0.54) (67,0.52) (68,0.52) (69,0.52) (70,0.52) (71,0.51) (72,0.50) (73,0.50) (74,0.48) (75,0.48) (76,0.47) (77,0.46) (78,0.46) (79,0.45) (80,0.43) (81,0.42) (82,0.41) (83,0.41) (84,0.40) (85,0.39) (86,0.39) (87,0.39) (88,0.38) (89,0.38) (90,0.38) (91,0.38) (92,0.36) (93,0.36) (94,0.35) (95,0.35) (96,0.35) (97,0.35) (98,0.33) (99,0.33) (100,0.33) (101,0.32) (102,0.30) (103,0.29) (104,0.28) (105,0.28) (106,0.26) (107,0.26) (108,0.25) (109,0.24) (110,0.23) (111,0.21) (112,0.20) (113,0.20) (114,0.20) (115,0.19) (116,0.19) (117,0.18) (118,0.17) (119,0.15) (120,0.13) (121,0.12) (122,0.11) (123,0.08) (124,0.07) (125,0.07) (126,0.06) (127,0.05)
    };
    
    \legend{Sparsely Gated MoE-1\textsuperscript{st}-Expert, Sparsely Gated  MoE-2\textsuperscript{nd}-Expert, Switch, BASE (training), BASE (testing)}
    
\end{axis}
\end{tikzpicture}
\caption{Expert Balancing in different Sparse Expert approaches across 128 experts, as measured on the validation set. Results for Sparsely Gated MoE and Switch are an average across all expert layers. BASE layers learn a reasonably balanced routing with no auxiliary balancing loss.}
\label{fig:expert_balance}
\end{figure}
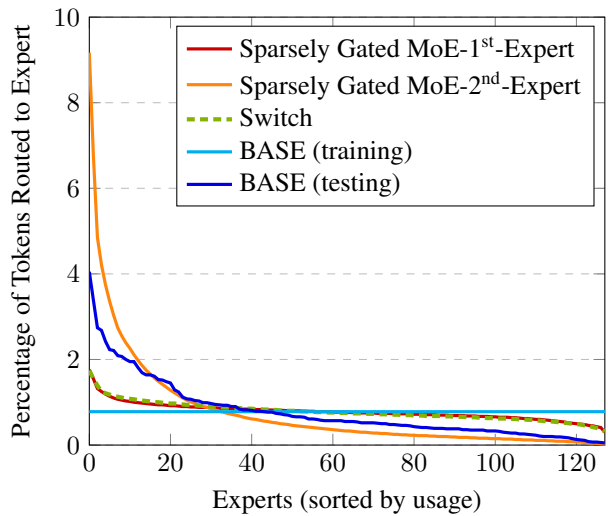

\subsection{Expert Balancing}
\label{section:balance}

A key difference between our model and other recent proposals is that we algorithmically balance token/expert assignments during training, instead of adding an additional loss function to balance assignments. However, both use greedy assignments at test time. 

We investigate whether our model learns a balanced assignment without an explicit balancing loss. \autoref{fig:expert_balance} shows the percentage of tokens assigned to each expert, sorted from most used to least used. Unsurprisingly, the top-1 assignment from BASE is less balanced than those from models with explicit balancing loss terms. However it is notably \emph{more} balanced than the 2\textsuperscript{nd} expert in the Sparsely Gated MoE model, and confirms that reasonably balanced assignments can be learnt without balancing losses. 

\subsection{Expert Specialization}
We also analyse how experts learn to specialize. Observing sample passages, we find that many assignment decisions appear to depend primarily on very local syntactic information. In particular, we found that the token input at timestep $t$ is often highly indicative of the expert assigned at time $t$.

Table \ref{table:experts} shows the most frequent previous input token when selected experts are chosen. 
We see clusters corresponding to quantities (5), numbers (42), possessives (125), subword fragments (101), and clusters of related verbs (72, 74, 126), nouns (23,27,36,43,76,84,96,98,105) and adjectives (9,81).
These tokens may tend to have similar distributions over next tokens.
This analysis suggests the model primarily assigns experts based on fairly superficial signals, and may motivate even simpler techniques for future work. 

\begin{table}[t]
\begin{tabular}{c|l}
\toprule
Expert & Top 5 Proceeding Tokens\\
\midrule
5 & \emph{year}, \emph{years}, \emph{billion}, \emph{million}, \emph{tonnes} \\
8 & \emph{people}, \emph{who}, \emph{Man}, \emph{everyone}, \emph{one} \\
9 & \emph{electronic}, \emph{local}, \emph{public}, \emph{national}, \emph{outdoor} \\
23 & \emph{funding}, \emph{budget}, \emph{benefits}, \emph{pressure}, \emph{price} \\
27 & \emph{Mustang}, \emph{State}, \emph{Center}, \emph{ation}, \emph{Grande} \\
34 & \emph{to}, \emph{will}, \emph{should}, \emph{it}, \emph{may} \\
36 & \emph{business}, \emph{bank}, \emph{financial}, \emph{science}, \emph{school} \\
42 & \emph{two}, \emph{50}, \emph{1}, \emph{80}, \emph{000} \\
43 & \emph{Bank}, \emph{Development}, \emph{.}, \emph{Construction}, \emph{Plant} \\
62 & \emph{work}, \emph{started}, \emph{involved}, \emph{working}, \emph{launched} \\
72 & \emph{is}, \emph{was}, \emph{be}, \emph{been}, \emph{were} \\
74 & \emph{going}, \emph{go}, \emph{come}, \emph{back}, \emph{return} \\
76 & \emph{painting}, \emph{case}, \emph{song}, \emph{statement}, \emph{discussion} \\
81 & \emph{new}, \emph{major}, \emph{bad}, \emph{larger}, \emph{grand} \\
84 & \emph{Ret}, \emph{Inspect}, \emph{Pl}, \emph{Pos}, \emph{Architect} \\
96 & \emph{US}, \emph{UNESCO}, \emph{government}, \emph{state}, \emph{UN} \\
98 & \emph{waiver}, \emph{procedures}, \emph{warrant}, \emph{status}, \emph{loans} \\
101 & \emph{B}, \emph{T}, \emph{W}, \emph{H}, \emph{k} \\
105 & \emph{app}, \emph{Windows}, \emph{Microsoft}, \emph{board}, \emph{10} \\
125 & \emph{his}, \emph{'s}, \emph{its}, \emph{their}, \emph{our} \\
126 & \emph{said}, \emph{says}, \emph{means}, \emph{noting}, \emph{out} \\
\bottomrule
\end{tabular}
\caption{Most frequent previous words for selected experts, showing that some experts assignment decisions are made based on very local contexts. 
For many other experts, the assignment decision depends on longer context, and is harder to visualize. 
}
\label{table:experts}
\end{table}

\subsection{Efficiency}
\begin{table}[t]
    \centering
    \begin{tabular}{l|c}
    \toprule
    Model & Tokens per Second \\
    \midrule
    Data Parallel        &  600k  \\
    Model Parallel $\times$2    &  224k  \\
    Sparsely Gated MoE   &  292k  \\
    Switch               &  469k  \\
    BASE                 &  545k  \\
    BASE $\times$2       &  475k  \\
    \bottomrule

    \end{tabular}
    \caption{Number of tokens processed per second during training by different models. BASE computes updates faster than other approaches that divide models over multiple workers, due to reduced communication overheads. This allows a 43B parameter model to be trained at 90\% of the speed of a 1.5B data parallel baseline.}
    \label{tab:efficiency}
\end{table}

While we focus on evaluating the compute efficiency of models, we note that there are substantial differences in the speed at which models process tokens. Table \ref{tab:efficiency} shows the number of tokens processed per second by different models during training, using 128 GPUs. Simple data parallel training is unsurprisingly the fastest, but BASE layers compute updates faster than other approaches due to reduced communication between workers. 
For the same compute efficiency, models which process tokens more slowly are more sample efficient, and may be preferable in lower data regimes.

\section{Related Work}

\citet{moe,gshard} introduce sparsely gated mixtures of experts layers, demonstrating how large sparse models can be trained efficiently by routing inputs to appropriate specialist workers. \citet{switch} show the design can be simplified by routing tokens to only a single worker. We further simplify the framework, by eliminating balancing loss functions, and showing the effectiveness of using only a single expert layer.

Sparse training is a line of work where traditional architectures are trained with sparse instead of dense layers and the number of parameters allowed during training is restricted to a percentage of the dense layers \cite{sparsemomentum, riggingthelottery, dynamicrepara}. Unlike our approach, these networks have fine-grained sparsity patterns which reduce overall FLOPS but make it difficult to achieve runtime benefits on modern accelerators like GPUs, which require contiguous memory segments for efficient processing. Since experts consist of sizable contiguous memory segments, our approach can utilize GPUs effectively.

Perhaps the most common use of sparse layers is in adding language-specific layers to machine-translation systems \cite{bapna2019simple,angela_mmmt}, or task-specific layers to pre-trained language models \cite{adapter}. Here, the expert assignment problem is hard coded, based on the task being solved or the language being translated. We instead explore learnable routing, which is applicable to problems where such structure is not available.

Other papers have explored alternative methods for adding very high capacity layers to neural language models.
For example, \citet{pkn} introduce a large memory layer that supports efficient sparse queries.
\citet{knnlm} show large gains from augmenting a language model with a nearest neighbour classifier over the training set, which recent work has also shown is applicable to machine translation \cite{knnmt}.

An orthogonal strand of work has improved the efficiency of transformer attention mechanisms, often by making them sparse \cite{child2019generating,correia2019adaptively,roy2020efficient}. We instead develop a sparse version of the other major component of the transformer, the feed forward network. 

\section{Conclusion}
We introduced a simple sparse BASE layer, which can be used to increase the capacity of any neural model, with little increase in training cost or complexity.
We demonstrate strong performance relative to both dense models and previously proposed sparse models.
Future work should explore more efficient implementations for computing balanced assignments, to further improve training speed.

\bibliography{main}
\bibliographystyle{icml2021}
\appendix
\begin{figure*}
    \centering
    \input{assignment_code}
    \caption{Algorithm used for solving linear assignment problem, adapted from 
    \citet{auction}.
    To mitigate the worst case performance, we switch to a greedy algorithm after \emph{max\_iterations}. While standard libraries are available for solving linear assignment problems, we found this algorithm more efficient for our use case. 
    }
    \label{fig:balanced_assignment}
\end{figure*}
\end{document}